\pdfoutput=1

\documentclass[11pt]{article}

\usepackage[preprint]{acl}

\usepackage{times}
\usepackage{latexsym}

\usepackage[T1]{fontenc}

\usepackage[utf8]{inputenc}

\usepackage{microtype}

\usepackage{inconsolata}

\usepackage{graphicx}
\usepackage{framed}

\usepackage{booktabs}
\usepackage{caption} 
\usepackage{multirow} 
\usepackage{tabularray}
\usepackage{tcolorbox}
\title{Conservative Bias in Large Language Models: Measuring Relation Predictions}

\author{
    Toyin Aguda,
    Erik Wilson,
    Allan Anzagira,
    Simerjot Kaur,
    Charese Smiley
    \\
  \texttt{\{toyin.d.aguda, erik.wilson, allan.anzagira, simerjot.kaur, charese.h.smiley\}@jpmchase.com}
}

\begin{document}
\maketitle
\begin{abstract}
Large language models (LLMs) exhibit pronounced {\it conservative bias} in relation extraction tasks, frequently defaulting to \textsc{no\_relation} label when an appropriate option is unavailable. While this behavior helps prevent incorrect relation assignments, our analysis reveals that it also leads to significant information loss when reasoning is not explicitly included in the output. We systematically evaluate this trade-off across multiple prompts, datasets, and relation types, introducing the concept of Hobson's choice to capture scenarios where models opt for safe but uninformative labels over hallucinated ones. Our findings suggest that conservative bias occurs twice as often as hallucination. To quantify this effect, we use SBERT and LLM prompts to capture the semantic similarity between conservative bias behaviors in  constrained prompts and labels generated from semi-constrained and open-ended prompts. 
\end{abstract}

\section{Introduction}

Recent advancements in LLMs have shown impressive ability to capture rich semantic knowledge and excel in tasks like text generation and question answering ~\cite{wadhwa-etal-2023-revisiting}. As these models are increasingly deployed for complex natural language processing tasks, including relation extraction, distinct behavioral patterns have emerged that warrant careful examination.

One such pattern is hallucination, where LLMs generate content (or relations) beyond the provided context (or available options). This phenomenon has attracted enormous attention within the LLM community ~\cite{sriramanan2024llmcheck, pmlr-v235-zhang24ay}, as it is often perceived as a limitation in most applications. However, hallucination also presents opportunities for innovation, particularly in domains 
 that benefit from creative generation such as image synthesis and other generative AI applications ~\cite{jiang2024surveylargelanguagemodel}.

Given the substantial research on hallucination detection ~\cite{yehuda-etal-2024-interrogatellm, li-etal-2024-dawn}, we have observed a reduction in hallucination rates. This reduction has led us to explore other emergent behaviors of LLMs that may have significant downstream effects. We focused on relation extraction tasks using LLMs, where we initially anticipated some degree of hallucination. However, our findings revealed minimal occurrences of such behavior. Instead, we observed a distinct pattern where LLMs consistently exhibit a systematic bias towards classifying instances as \textsc{no\_relation} even when a more appropriate relation is available, presenting a "Hobson's choice" scenario. We attribute this behavior to alignment strategies designed to reduce hallucinations by reinforcing contextual adherence while suggesting external alternatives. We define this preference for overly cautious responses as \textbf{\textit{Conservative Bias (CB)}}. 

In relation extraction tasks, LLMs exhibit a distinct CB, defaulting to the least incorrect classification when faced with uncertainty. Unlike hallucination, this bias leads to a unique form of information loss by creating "Hobson's choice" scenarios, where models favor safe but uninformative labels even when more suitable alternatives exist.

\begin{figure}[h!]
    \centering
    \includegraphics[width=0.5\textwidth]{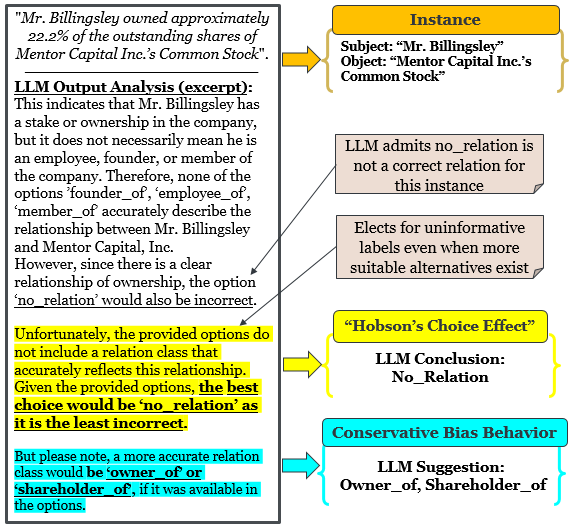}
    \caption{Example LLM Output from REFinD dataset demonstrating Hobson's Choice and Conservative Bias behavior.}
    \label{fig:CBB-Out}
\end{figure}

Our Figure \ref{fig:CBB-Out} example shows a LLM's response to relation extraction task, especially when the available options do not perfectly match the true relationship. The example demonstrates the model's strategies in "Hobson's choice" scenarios and conservative bias (CB) behavior. The "Hobson's choice" effect occurs when the LLM must select from predefined options that do not accurately capture the relationship, defaulting to \textsc{no\_relation} as the least incorrect choice. The LLM output also shows its conservative bias behavior, acknowledging that a more accurate relation like \textsc{owner\_of} or \textsc{shareholder\_of} would be preferable if available. This conservative approach helps avoid incorrect assertions but can lead to a loss of valuable information.

This work seeks to address three key research questions: [RQ1] How can we capture and quantify this CB? [RQ2] What is its relationship to hallucination prevention? [RQ3] How can we leverage this behavior to improve relation extraction tasks?


\section{Related Work}
Recent studies have highlighted emergent behaviors in LLM, such as sycophancy and hallucination, which impact their reliability and trustworthiness in downstream applications ~\cite{rrv-etal-2024-chaos}. Sycophancy refers to the tendency of models to align their responses with user views, regardless of objective correctness \cite{sharma2025understandingsycophancylanguagemodels}. This behavior is most prevalent in models that have been fine-tuned using human feedback, but it can be mitigated through the use of synthetic data \cite{wei2024simplesyntheticdatareduces}. 

Hallucinations have received much attention from the research community \cite{10.1145/3703155, sahoo2024comprehensive,Huang_2025}. Among proposed hallucination mitigation methods, \citet{su2024mitigatingentitylevelhallucinationlarge}
 investigated LLM hallucinations in entity/relation extraction tasks proposing mitigating techniques. Advances in prompt engineering ~\cite{wadhwa-etal-2023-revisiting} have also mitigated hallucinations by constraining responses to given contexts ~\cite{sadat-etal-2023-delucionqa}.

As prompt engineering advances, new emergent behaviors in LLMs may arise. To the best of our knowledge, Conservative Bias behavior has not been explored in existing literature.

\section{Method}

Our research aims to analyze CB in LLMs. We investigate the frequency with which LLMs default to the least incorrect labels from a list of options, as opposed to generating hallucinated relations. We analyze the rates of hallucination and CB across multiple prompt iterations. We also explore practical applications where CB can be utilized to refine relation classification, potentially expanding existing relations.

Formally, CB is detected in an output when the following conditions are met: \textbf{(i.)} the model recognizes that a valid relation exists, as outputted in the reasoning. \textbf{(ii.)} the correct relation type is not available in the option set. \textbf{(iii.)} the model chooses to default to \textsc{no\_relation} or selects the least incorrect (suboptimal) option. \textbf{(iv.)} the model demonstrates awareness of the correct relation through reasoning, suggesting it when appropriate to preserve the integrity of extracted relations. See the example in Figure~\ref{fig:CBB-Out}. 
 
For evaluation purposes, we designed three types of prompts: Constrained Prompt, Semi-constrained Prompt, and Open-ended Prompt and assessed performance using four measures: Hobson's Choice Rate ({\bf HCR}), Conservative Bias Rate ({\bf CBR}), Hallucination Rate ({\bf HR}) and New Relation Rate ({\bf NRR}).

\subsection{Prompting Design}
\begin{figure}[h]
    \centering
    \includegraphics[width=0.40\textwidth]{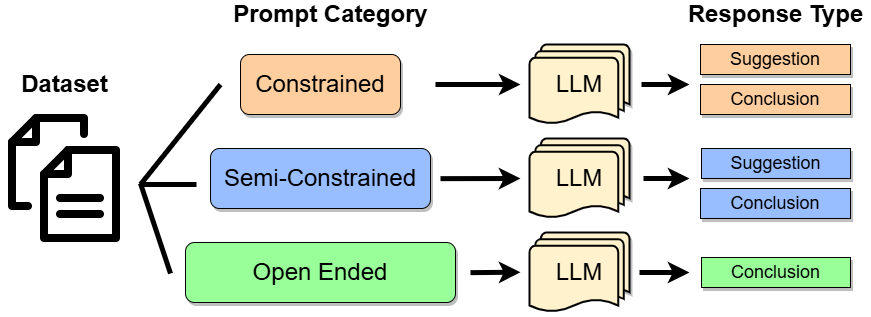}
    \caption{Process Workflow.}
    \label{fig:enter-label}
\end{figure}
We adopted a multi-tiered approach to prompt design, where each level offers varying degrees of specificity to the LLMs. This approach explores how different levels of constraint affect the LLMs' ability to generate and select appropriate relations. 
The prompt categories are defined as follows:

\textit{Open-ended Prompts}: represent the least constrained interaction with LLMs. In this setup, no predefined list of relation classes is provided. Instead, the LLMs are tasked with generating the most suitable relation between subject and object based on the input data. 

\textit{Semi-Constrained Prompts}: offer a moderate level of guidance. Here, LLMs are provided with a list of relations to choose from, which varies based on entity-pair type. However, the models retain the flexibility to propose a relation if none of the provided options are deemed most appropriate.

\textit{Constrained Prompts}: are the most restrictive, requiring LLMs to select the best relation from a predefined list of options (relation classes). These prompts are designed to assess the LLMs' judgment and decision-making capabilities when faced with a limited set of possibilities.

By employing this tiered prompting strategy, as seen in Figure~\ref{fig:enter-label}, we provide the LLMs with multiple perspectives before prompting them to select a final label, aiming to provide enough context for detecting relations between subject and object that might be missed by a human labeler. More detailed examples of these prompts can be found in appendix \ref{app:detailed-prompts}.

\subsection{Metrics}

HCR metric captures how often the model defaults to \textsc{no\_relation}  (or suboptimal option) due to the lack of a perfectly suitable option among the provided choices. CBR metric quantifies how often the LLM suggests a more suitable relation in its reasoning, even though it ultimately defaults to \textsc{no\_relation} as the least incorrect option. This highlights the model's tendency to prioritize avoiding explicit mistakes over making informative predictions.

\begin{equation}
    HCR = \frac{N^{HC}}{N^{total}},\,\, CBR = \frac{N^{CB}}{N^{HC}}
\end{equation}
where  \( N^{HC} \) = Number of times the model selects \textsc{no\_relation} as the least incorrect option; \( N^{total} \) = Total number of relation extraction tasks and 
 \( N^{CB} \) = Number of times where model exhibits CB, meaning the model suggests a more suitable relation in its reasoning, but opts for \textsc{no\_relation} in order to preserve LLM accuracy. The key distinction between \( N^{HC} \) and \( N^{CB} \) lies in the model's recognition of a valid relation which can only be extracted from the reasoning output.

The HR quantifies how often the LLM generates an unsupported or non-existent relation, primarily captured in \textit{constrained} prompt, while the NRR measures how often the LLM proposes a valid relation not present in the provided options, as captured in \textit{semi\_constrained} and \textit{open\_ended} prompts, thereby justifying the correctness of conservative bias behavior (CB).
\begin{equation}
    HR = \frac{N^{H}}{N^{total}},\,\, NRR = \frac{N^{NR}}{N^{total}}
\end{equation}
where \( N^{H} \) = Number of times the model hallucinates (i.e., generates a relation that is factually incorrect or not supported by the input data), where \( N^{NR} \) = Number of times the model suggests a valid relation that is not present in the predefined option set. \(N^{total} \) = Total number of relation extraction tasks.

A major distinction between Hallucination and Conservative Bias Behavior in Constrained Prompt outputs is that Hallucination is a manifestation of incorrect information as LLM output, while CB involves recognition of a better alternative option from reasoning despite a conclusive conservative choice. See table \ref{tab:tradeoff-tab} in the appendix for more details.

\section{Experiments and Results}
\paragraph{Data}

For our experiment, we focus on two datasets: REFinD \cite{10.1145/3539618.3591911} and TACRED \cite{zhang2017position}.\footnote{Dataset statistics can be obtained in their original papers.} 
REFinD is a large financial dataset, consisting of 28,676 instances and 22 relation types across 8 entity pairs, with data sourced from the quarterly and annual reports of publicly traded companies. TACRED, is a large-scale relation extraction dataset with 106,264 examples derived from newswire and web text and features 41 relation types. 
For our analysis, we focus on subset of data where gold\_relation is labeled as \textsc{no/other relation} or \textsc{no\_relation} which constitutes 45\% of the REFinD and 79.5\% of the TACRED dataset (statistics shown in App \ref{app:datasetstats}). We also explored, but did not include, BioRED \cite{Luo_2022}, a biomedical relation extraction dataset, because this dataset has no \textsc{no\_relation} option.

\paragraph{Models}
We leveraged GPT-4, Llama3.1-8B-Instruct, and Mistral-7B-Instruct-v0.3 as our models. For each model, we utilized two low temperature settings, specifically 0.2 and 0.5, in order to simulate multiple data annotators while maintaining high level of determinism. We captured output consistency per temperature by conducting multiple iterations of each prompt. Our selection of LLM models was based on those accessible within our organization that are on the LLM leaderboard.

\paragraph{Prompt Setup}
Prompts are structured in a hierarchical manner, allowing us to evaluate how varying level of constraints can affect LLMs' responses. While all prompts share the same basic structure, they differ in their option list setup. Among these, only the \textit{constrained} prompt is prone to hallucination. Outputs from \textit{semi-constrained} and \textit{open-ended} prompts will be used to validate the CB behavior in the \textit{constrained} prompt.

\subsection{Results}

\subsubsection{Model Performance}
Across different temperature settings and prompt configurations, we observe a range of outcomes when using the ``step-by-step'' instruction~\cite{lightman2023letsverifystepstep}. We analyze the outputs and categorize responses into `conclusions' and `suggestions' for both constrained and semi-constrained prompt responses. The results reveal distinct patterns in hallucination mitigation and the manifestation of CB. 

On the REFinD dataset, GPT-4 outperformed Llama3.1, exhibiting a notably low HR of 0.02-0.04\% and CBR of 1-1.33\% for the constrained prompt. This pattern persists with the semi-constrained prompt, where we observe NRR of 7-10\% and CBR of 37-41\%. 

Our analysis, summarized in Table~\ref{tab:metric}, shows that the CBR can be more prevalent than the HR in relation extraction tasks. While GPT-4 demonstrated strong hallucination resistance under constrained prompting, the transition to semi-constrained prompt yielded interesting dynamics: although models showed an increased tendency to suggest novel relations when explicitly allowed, we observed a concurrent quadrupling (4x) of the CBR compared to NRR. In the semi-constrained scenarios, GPT-4 frequently generated novel relation suggestions but exhibited reluctance in conclusively asserting them (avoiding hallucinations), often defaulting to \textsc{no\_relation} or selecting from other predefined options. 

To further substantiate our findings, we conducted additional tests on a data subset utilizing the Mistral-7B-Instruct-v0.3 model. The constrained prompts results for the Tacred dataset mirrored those observed with Llama3.1, demonstrating elevated Conservative Bias rate (CBR) alongside reduced Hallucination Rates (HR). The output from REFinD exhibited sensitivity to temperature settings. At a temperature of 0.2, the HR was marginally higher than the CBR, whereas at a temperature of 0.5, a significantly higher CBR was observed, accompanied by a low HR. This underscores the anticipated inverse correlation between CBR and HR [RQ2].

\begin{table}[h!]
\centering
\resizebox{1.0\columnwidth}{!}{
\begin{tabular}{l|c|c|c|c|c|c}
\specialrule{1.5pt}{0pt}{0pt}
\textbf{Prompt} & \textbf{Dataset} & \textbf{Temp} & \textbf{CBR\%} & \textbf{HR\%} & \textbf{NRR\%} & \textbf{HCR\%} \\
\specialrule{1.5pt}{0pt}{0pt}
\multicolumn{7}{c}{\textbf{GPT-4}}\\
\specialrule{1.5pt}{0pt}{0pt}
\multirow{4}{*}{Const.} & \multirow{2}{*}{REFinD} & 0.2  & 1.14 & 0.04 & - & 57.72 \\
   &        & 0.5  & 1.33 & 0.06 & - & 64.37 \\
  \cline{2-7}
   & \multirow{2}{*}{TACRED} & 0.2  & 7.99 & 15.47 & - & 1.23 \\  
   &        & 0.5  & 7.11 & 13.87 & - & 4.86 \\ 
\hline
\multirow{4}{*}{Semi} & \multirow{2}{*}{REFinD} & 0.2  & 37.67 & - & 9.75 & 69.15 \\
   &        & 0.5  & 40.68 & - & 7.27 & 67.29 \\
  \cline{2-7}
  & \multirow{2}{*}{TACRED} & 0.2 & 9.70 & - & 28.08 & 2.80 \\
   &        & 0.5  & 9.20 & - & 27.04 & 10.46 \\
   \hline
   \multirow{4}{*}{Open} & \multirow{2}{*}{REFinD} & 0.2  & - & - & 81.66 & - \\
   &        & 0.5  & - & - & 81.78 & - \\
  \cline{2-7}
  & \multirow{2}{*}{TACRED} & 0.2 & - & - & 82.46 & - \\
   &        & 0.5  & - & - & 76.96 & - \\
\specialrule{1.5pt}{0pt}{0pt}
\multicolumn{7}{c}{\textbf{Llama3.1}}\\
\specialrule{1.5pt}{0pt}{0pt}
\multirow{4}{*}{Const.} & \multirow{2}{*}{REFinD} & 0.2  & 0.29 & 8.18 & - & 2.63  \\
   &        & 0.5  & 1.07 & 4.67 & - & 1.44 \\
   \cline{2-7}
  & \multirow{2}{*}{TACRED} & 0.2 & 10.80 & 9.60 & - & 75.00 \\
   &        & 0.5  & 4.00 & 2.60 & - & 100.00 \\
   \specialrule{1.5pt}{0pt}{0pt}
\multicolumn{7}{c}{\textbf{Mistral}}\\
\specialrule{1.5pt}{0pt}{0pt}
\multirow{4}{*}{Const.} & \multirow{2}{*}{REFinD} & 0.2  & 2.51 & 3.84 & - & 57.14  \\
   &        & 0.5  & 19.15 & 2.97 & - & 12.03 \\
   \cline{2-7}
  & \multirow{2}{*}{TACRED} & 0.2 & 13.01 & 8.65 & - & 29.53 \\
   &   & 0.5  & 15.57 & 7.83 & - & 32.57 \\
   \specialrule{1.5pt}{0pt}{0pt}
\end{tabular}
}
\caption{LLM Outputs by Prompt Type. "-" denotes N\//A.}
\label{tab:metric}
\end{table}

To assess the semantic validity of CB labels identified in the constrained prompt, we conducted a semantic similarity analysis using outputs from semi-constrained and open-ended prompts. Focusing on instances flagged for CB, we found that over 57\% of CB-flagged instances in the REFinD dataset defaulted to Hobson's choice, as detailed in Table~\ref{tab:metric}. These findings provide quantitative insights into the detection and measurement of CB in LLMs, addressing our primary research question regarding the characterization and quantification of this CB phenomenon [RQ1].

\subsubsection{Quality of LLM-Generated Relations}

To evaluate the semantic quality of LLM-generated relations, we employ two methods: SBERT (Sentence-BERT) and a semantic similarity prompt executed via GPT-4. All semantic similarity scores range from 0 to 1. We set our similarity threshold to 0.7 to align with established benchmarks~\cite{okazaki-tsujii-2010-simple}. Notably, the GPT-4  prompt consistently yielded a higher proportion for scores above the threshold of 0.7, when compared to SBERT. 

As shown in Table ~\ref{tab:semantic_similarity} and the snippet below in Table ~\ref{tab:snippet_semantic_similarity}, results for the REFinD dataset indicate that at temperatures of 0.2 and 0.5, the CB captured in the constrained prompt was similar in meaning to the LLM output for the same instance in the semi-constrained prompt by 62\% and 59\%, respectively. When compared to the open-ended prompt, the similarity was 59\% and 54\%, respectively. In contrast, the TACRED dataset exhibited much lower percentages. These findings suggest that the semantic alignment between constrained and semi-constrained outputs is more robust in the REFinD dataset. Consequently, we can leverage CB labels as plausible conclusions to improve relation extraction tasks [RQ3].

\begin{table}[h!]
\centering
\resizebox{1.0\columnwidth}{!}{
\begin{tabular}{l c c c c c c c}
\toprule 
\multirow{2}{*}{\textbf{Prompt}} & 
\multirow{2}{*}{\textbf{\parbox[c]{2cm}{\centering Semantic\\Similarity}}} &
\multicolumn{2}{c}{\textbf{REFinD Semi}} &
\multicolumn{2}{c}{\textbf{TACRED Semi}}  \cr
 \cmidrule(lr){3-4} \cmidrule(lr){5-6} \cmidrule(lr){7-8}
  &  &
 \textbf{$> $0.7} & \textbf{$\mu$} &
 \textbf{$>$0.7} & \textbf{$\mu$} \cr
\midrule
\multirow{2}{*}{\parbox[c]{1.8cm}{Const.\\Temp: 0.2}} & SBERT 
 & 34\% & $0.54_{\pm 0.30}$ & 4\% & $0.30_{\pm 0.22}$ \\
 & GPT-4 Prompt
 & 62\% & $0.44_{\pm 0.35}$ & 11\% & $0.35_{\pm 0.26}$ \\
\midrule
 \multirow{2}{*}{\parbox[c]{1.8cm}{Const.\\Temp: 0.5}} & SBERT 
 & 41\% & $0.55_{\pm 0.33}$ & 5\% & $0.25_{\pm 0.22}$ \\
 & GPT-4 Prompt 
 & 59\% & $0.65_{\pm 0.36}$ & 8\% & $0.30_{\pm 0.24}$ \\
\bottomrule
\end{tabular}
}
\caption{Snippet: Semantic Similarity scores for REFinD and TACRED from GPT-4 output.}
\label{tab:snippet_semantic_similarity}
\end{table}

We calculated the inter-annotator agreement (IAA) to assess reliability, using Cohen's Kappa ($\kappa$) across multiple model runs at consistent temperature settings. Our results demonstrated substantial agreement ($\kappa$ = 0.65-0.80), indicating significant reliability~\cite{mchugh2012interraterreliability} of our generated relations (particularly for GPT-4 relation extraction). Both Spearman's rank correlation and Cohen's Kappa lead to the same conclusion: higher reliability for GPT-4 relation extractions and lower for Llama3.1.

\section{Discussion}
Our findings confirm the presence of CB tendencies in LLMs during relation extraction. While GPT-4 demonstrates strong hallucination resistance under constrained conditions (0.02-0.04\% HR), it also shows a much higher frequency of conservatism. This pattern persists in the semi-constrained design, suggesting a fundamental tension between innovation and accuracy in LLM behavior. In contrast, Llama3.1 shows less CB but a higher HR. This indicates that as models become more resistant to hallucinations, they tend to exhibit increased CB [RQ2], presenting a crucial trade-off in model behavior that requires careful consideration in application design. 

There are significant differences in output quality between GPT-4 and Llama3.1 when using identical prompts. Llama3.1 generated noisier outputs, often returning meta-responses such as \textit{``Please specify title example''}, resulting in substantial data loss during the cleanup process. This disparity in output quality highlights the importance of model selection and prompt engineering in relation extraction tasks. To mitigate this limitation, our research indicates that detailed prompting strategies incorporating step-by-step reasoning are essential. This finding is particularly relevant in specialized professional contexts; for example - \textit{A boutique law firm employing AI for litigation analysis}. Without structured reasoning steps in the prompting strategy, these systems risk returning conclusions that may be either overly conservative or inappropriately broad, potentially missing crucial legal nuances within the established constraints.

\section{Conclusion \& Future Work}
This study explored the Conservative Bias (CB) in LLMs during relation extraction, where models default to \textsc{no\_relation} when a correct option is unavailable. Our experiment confirmed an inverse relationship between Conservative Bias Rate (CBR) and Hallucination Rate (HR), highlighting a trade-off between accuracy and innovation. Future research should focus on developing prompting strategies that balance CB with the need for novel relation identification, potentially by refining prompt designs and integrating external knowledge bases. 

\section{Disclaimer}
This paper was prepared for informational purposes by the Artificial Intelligence Research group of JPMorgan Chase \& Co. and its affiliates ("JP Morgan'') and is not a product of the Research Department of JP Morgan. JP Morgan makes no representation and warranty whatsoever and disclaims all liability, for the completeness, accuracy or reliability of the information contained herein. This document is not intended as investment research or investment advice, or a recommendation, offer or solicitation for the purchase or sale of any security, financial instrument, financial product or service, or to be used in any way for evaluating the merits of participating in any transaction, and shall not constitute a solicitation under any jurisdiction or to any person, if such solicitation under such jurisdiction or to such person would be unlawful.


\section{Limitations}
While our work provides novel insights into CB detection in relation extraction tasks using LLMs, we acknowledge some limitations. We focused majorly on two LLMs (GPT-4, and Llama3.1-8B-Instruct) and tested a subset of the dataset on Mistral-7B-Instruct-v0.3, limiting the generalizability of our findings across other models. Although we introduced metrics to quantify CB occurrences, there is a need for more robust evaluation frameworks to capture nuanced aspects of CB. Additionally, we noticed that the quality of datasets used can significantly impact the results.

 The study primarily relied on automated metrics for evaluation. Incorporating human evaluation could provide a more nuanced understanding of the quality and relevance of the extracted relations. Finally, as LLMs and their training data evolve, the behavior of models regarding CB and hallucination might change. The findings may need to be revisited with newer versions of models and updated datasets. 
 
 As this work represents one of the first systematic investigations of Conservative Bias (CB) in relation extraction, our findings should be considered initial benchmarks rather than definitive measurements. We hope this paper will spur further research into CB detection and mitigation strategies in LLMs, extending beyond relation extraction tasks.

\section{Ethics Statement}
This research was conducted with a focus on ethical standards, particularly in addressing the CB in LLMs for relation extraction tasks. We used publicly available datasets, REFinD and TACRED, acknowledging potential biases inherent in them. Our study does not involve human subjects or personal data, minimizing privacy concerns. Our findings serve as initial benchmarks, and we encourage further research to explore ethical implications and enhance the social benefits of LLMs.


\nocite{Ando2005,andrew2007scalable,rasooli-tetrault-2015}

\onecolumn
\clearpage \appendix
\section{Appendix}
\renewcommand\thefigure{\thesection.\arabic{figure}}
\renewcommand\thetable{\thesection.\arabic{table}}

\subsection{Semantic Similarity Scores} 
\label{app:simscores}

\begin{figure}[h!]
    \centering
    \captionsetup{font=small}
    \includegraphics[width=0.50\textwidth]{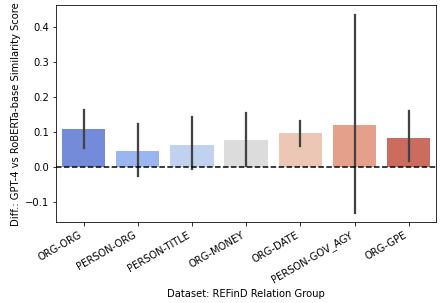}
    \caption{REFinD: Difference in Semantic Similarity Scores (GPT4 vs SBERT).}
    \label{fig:REFinD}
\end{figure}

\begin{figure}[h!]
    \centering
    \captionsetup{font=small}
    \includegraphics[width=0.52\textwidth]{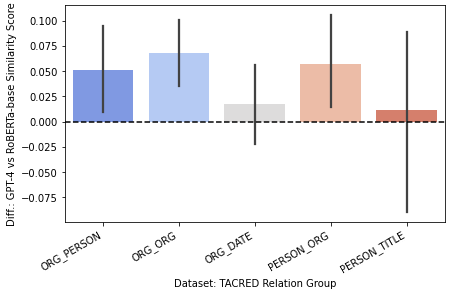}
    \caption{TACRED: Difference in Semantic Similarity Scores (GPT4 vs SBERT).}
    \label{fig:TACRED}
\end{figure}

\begin{table*}[h!]
\centering
\resizebox{\linewidth}{!}{
\begin{tabular}{l c c c c c c c c c}
\toprule 
\multirow{2}{*}{\textbf{Description}} & 
\multirow{2}{*}{\textbf{\parbox[c]{2cm}{\centering Semantic\\Similarity}}} &
\multicolumn{2}{c}{\textbf{REFinD Semi}} &
\multicolumn{2}{c}{\textbf{REFinD Open}} &
\multicolumn{2}{c}{\textbf{TACRED Semi}} &
\multicolumn{2}{c}{\textbf{Tacred Open}} \cr
 \cmidrule(lr){3-4} \cmidrule(lr){5-6} \cmidrule(lr){7-8} \cmidrule(lr){9-10}
  &  &
 \textbf{$> $0.7} & \textbf{$\mu$} &
 \textbf{$> $0.7} & \textbf{$\mu$} &
 \textbf{$> $0.7} & \textbf{$\mu$} &
 \textbf{$>$0.7} & \textbf{$\mu$} \cr
\midrule
\multirow{2}{*}{\parbox[c]{4cm}{Constrained Prompt\\Temp - 0.2}} & SBERT 
 & 34\% & $0.54_{\pm 0.30}$ & 21\% & $0.46_{\pm 0.25}$ & 4\% & $0.30_{\pm 0.22}$ & 5\% & $0.26_{\pm 0.22}$ \\
 & GPT-4 Prompt
 & 62\% & $0.44_{\pm 0.35}$ & 59\% & $0.71_{\pm 0.22}$ & 11\% & $0.35_{\pm 0.26}$ & 10\% & $0.31_{\pm 0.25}$ \\
\midrule
 \multirow{2}{*}{\parbox[c]{4cm}{Constrained Prompt\\Temp - 0.5}} & SBERT 
 & 41\% & $0.55_{\pm 0.33}$ & 18\% & $0.45_{\pm 0.25}$ & 5\% & $0.25_{\pm 0.22}$ & 3\% & $0.24_{\pm 0.20}$\\
 & GPT-4 Prompt 
 & 59\% & $0.65_{\pm 0.36}$ & 54\% & $0.68_{\pm 0.24}$ & 8\% & $0.30_{\pm 0.24}$ & 11\% & $0.31_{\pm 0.25}$ \\
\bottomrule
\end{tabular}
}
\caption{Semantic Similarity scores for REFinD and TACRED from GPT-4 output.}
\label{tab:semantic_similarity}
\end{table*}

\begin{table}[!htbp]
\centering
\small
\captionsetup{font=small}
\begin{tabular}{c|c|c|c}
\specialrule{1.5pt}{0pt}{0pt}
\textbf{Model} & \textbf{Dataset} & \textbf{$\kappa$} & \textbf{$\rho$} \\ \specialrule{1.5pt}{0pt}{0pt}
\multirow{2}{*}{GPT-4} & REFinD & 0.65-0.77 & 0.66-0.79 \\ \cline{2-4}
                       & TACRED & 0.30-0.53 & 0.33-0.54 \\ \hline
Llama3.1 & REFinD & 0.31-1.0 & 0.32-1.0 \\ 
\specialrule{1.5pt}{0pt}{0pt}
\end{tabular}
\caption{Inter Annotator Agreement Scores. Metrics: Cohen's Kappa ($\kappa$) and Spearman's Correlation Coefficient ($\rho$) on dataset per model for multiple runs.}
\label{tab:prompt}
\end{table}

\subsection{Dataset Statistics} \label{app:datasetstats}

\begin{table}[h!]
\centering
\begin{tabular}{lllll}
\hline
Dataset   & Train & Dev  & Test & Total \\ \hline
REFinD    & 9128 & 1965 & 1953 & \textbf{13046} \\
TACRED & 55112  & 17195 & \textbf{12184} & 84491 \\ \hline
\end{tabular}
\caption{Datasets: No\_Relation set}
\label{tab:dataset}
\end{table}

\subsection{Detailed Prompt Breakdowns}
\label{app:detailed-prompts}
We've highlighted key distinctions between the prompts in bold below. Sections of the prompt denoted between \{braces\} were programmatically filled in based on the given example from the dataset.

\begin{center}
    \begin{tcolorbox}[title=Constrained Prompt:, colframe=black, colback=white, coltitle=black, fonttitle=\bfseries, colbacktitle=lightgray]
        Given the following sentence: \{highlighted\_text\}, with marked entities at specific positions as highlighted phrases: \{subject\} and \{object\}. Identify the relation between \{subject\} (positioned between the entity marker [SUBJ] \& [/SUBJ] tag) and \{object\} (positioned between the entity marker [OBJ] \& [/OBJ] tag) in the given sentence \{highlighted\_text\}. \textbf{Please choose the appropriate relation class from the following: \{options\} which best describes the relation between \{subject\} and \{object\}. If there is no relation between the marked entities in this sentence: {highlighted\_text}, then return 'no\_relation'.} Ensure the direction of the relation from the subject entity to object entity is considered and return the appropriate relation you selected from the relation class: \{options\}. To determine the appropriate relation class, let's think step by step. 
    \end{tcolorbox}
\end{center}

\begin{center}
     \begin{tcolorbox}[title=Semi-Constrained Prompt:, colframe=black, colback=white, coltitle=black, fonttitle=\bfseries, colbacktitle=lightgray]
         Given the following sentence: \{highlighted\_text\}, with marked entities at specific positions as highlighted phrases: \{subject\} and \{object\}. Identify the relation between \{subject\} (positioned between the entity marker [SUBJ] \& [/SUBJ] tag) and \{object\} (positioned between the entity marker [OBJ] \& [/OBJ] tag) in the given sentence \{highlighted\_text\}. \textbf{If the relation class is not part of the listed options: \{options\}, then provide the most appropriate relation class you can determine or come up with. Please choose the appropriate relation class from the following: \{options\} or suggest a relation which best describes the relationship between \{subject\} and \{object\}. If there is no relation between the marked entities in this sentence: \{highlighted\_text\}, then return 'no\_relation'.} Ensure the direction of the relation from the subject entity to object entity is considered, then return the appropriate relation and if you do not know return 'dont\_know'. Here are the relation options again: \{options\}. To determine the appropriate relation class, let's think step by step. 
    \end{tcolorbox}
\end{center}

\begin{center}
     \begin{tcolorbox}[title=Open Ended Prompt:, colframe=black, colback=white, coltitle=black, fonttitle=\bfseries, colbacktitle=lightgray]
        Given the following sentence: \{highlighted\_text\}, with marked entities at specific positions as highlighted phrases: \{subject\} and \{object\}; identify what the relation between \{subject\} (positioned between the entity marker [SUBJ] \& [/SUBJ] tag) and \{object\} (positioned between the entity marker [OBJ] \& [/OBJ] tag) is in the given sentence \{highlighted\_text\}. \textbf{That is, provide the most appropriate relation class you can suggest to capture the relation by \{subject\} and \{object\} in this sentence. If there is no relation between the marked entities in this sentence: \{highlighted\_text\}, then return 'no\_relation' or if you do not know just return 'dont\_know'.} Ensure the direction of the relation from the subject entity to object entity is considered and return the appropriate relation you can suggest that best captures this relation. To identify and suggest a relation class, let's think step by step.  
    \end{tcolorbox}
\end{center}

\clearpage

\subsection{Tradeoffs between Hallucination and Conservative Bias}
\label{app:tradeoffs}
\begin{table}[h!]
\centering
\begin{tblr}{
  width = \linewidth,
  colspec = {Q[194]Q[333]Q[413]},
  row{1} = {c},
  row{2} = {c},
  cell{1}{1} = {c=3}{0.94\linewidth},
  hlines,
  vlines,
}
\textbf{Trade-Off: Hallucination vs. Conservative Bias Behavior}~ ~ &  & \\
\textbf{Criteria}~ & \textbf{LLM Output as Hallucination}~ & \textbf{Conservative Bias Behavior}~\\
\textbf{Definition}~ & LLM generates new relation outside the option list provided as conclusion.~ & LLM concludes the least incorrect option (Hobson's Choice), often “No\_Relation,” from the finite list given while suggesting a more appropriate relation in the reasoning.~\\
\end{tblr}
\caption{Discussion of tradeoffs between Hallucination and Conservative Bias Behavior}
\label{tab:tradeoff-tab}
\end{table}

\subsection{Further Results for Llama3.1 and Mistral}
\label{app:llama}

\begin{table}[h!]
\centering
\resizebox{0.5\columnwidth}{!}{
\begin{tabular}{l|c|c|c|c|c|c}
\specialrule{1.5pt}{0pt}{0pt}
\textbf{Prompt} & \textbf{Dataset} & \textbf{Temp} & \textbf{CBR\%} & \textbf{HR\%} & \textbf{NRR\%} & \textbf{HCR\%} \\
\specialrule{1.5pt}{0pt}{0pt}
\multicolumn{7}{c}{\textbf{Llama3.1}}\\
\specialrule{1.5pt}{0pt}{0pt}
\multirow{2}{*}{Semi} & \multirow{2}{*}{REFinD} & 0.2  & 0.61 & - & 7.89 & 5.06 \\ 
   &        & 0.5  & 3.78 & - & 10.83 & 4.26 \\
   \hline
   \multirow{2}{*}{Open} & \multirow{2}{*}{REFinD} & 0.2  & - & - & 66.19 & - \\ 
   &        & 0.5  & - & - & 76.81 & - \\
   \specialrule{1.5pt}{0pt}{0pt}
\multicolumn{7}{c}{\textbf{Mistral}}\\
\specialrule{1.5pt}{0pt}{0pt}
\multirow{2}{*}{Semi} & \multirow{2}{*}{REFinD} & 0.2  & 16.39 & - & 5.58 & 16.25 \\ 
   &        & 0.5  & 19.20 & - & 7.12 & 16.00 \\
   \specialrule{1.5pt}{0pt}{0pt}
\end{tabular}
}
\caption{Further results for Llama and Mistral on Semi-Constrained and Open Ended Prompts.}
\label{tab:llama-extra}
\end{table}

\end{document}